\documentclass[runningheads,a4paper]{llncs}

\usepackage{amssymb}
\usepackage{graphicx}
\usepackage{cite}

\usepackage{url}
\urldef{\mailsa}\path|kmrocki@us.ibm.com|

\begin{document}

\mainmatter  % start of an individual contribution

\title{Towards Machine Intelligence}

\titlerunning{Towards Machine Intelligence}

\author{Kamil Rocki}
\authorrunning{Kamil Rocki}

\institute{IBM Research, San Jose, CA 95120, USA\\
\mailsa\\}


\toctitle{Lecture Notes in Computer Science}
\tocauthor{Authors' Instructions}
\maketitle

\begin{abstract}
There exists a theory of a single general-purpose learning algorithm which could explain the principles of its operation. This theory assumes that the brain has some initial rough architecture, a small library of simple innate circuits which are prewired at birth and proposes that all significant mental algorithms can be learned. Given current understanding and observations, this paper reviews and lists the ingredients of such an algorithm from both architectural and functional perspectives.
\end{abstract}

\section{Introduction}

Recently, much progress has been made in the area of supervised learning\cite{mohamed2009, Graves:09tpami, NIPS2012_4824, 1409.4842, graves2014,Mnih2015}{}. However, one of the greatest challenges remaining in artificial intelligence research is advancing the field of unsupervised learning algorithms\cite{DBLP:journals/corr/abs-1206-5538, Bengio-2009, lecuncvpr, Goodfellow-et-al-2015-Book}{}. Especially, autonomous learning of complex spatiotemporal patterns poses a great challenge.
This paper reviews and lists the ingredients of a possible general-purpose learning algorithm given current state of knowledge.

The neocortex, which is found only in mammals, is deemed to be the place where intelligence originates. It has been studied extensively over the past decades, but to date there is still no consensus on the principles of its operation. Some theories suggest that a single learning algorithm might be sufficient to explain intelligence\cite{citeulike:13329708, Hawkins:2004:INT:993636, kurzweil2012create, HintonSejnowski:86, domingos2015master}{}. Such theories have been considered ever since Mountcastle's discovery of the simple uniform architecture of the cortex\cite{Mountcastle:1978}{} (six horizontal layers organized into vertical structures called cortical columns; these columns can be thought of as the basic repeating functional units of the neocortex). This discovery might suggest that all brain regions perform similar operations, and there are no region-specific algorithms. Another famous experiment supporting this hypothesis showed that after rewiring, the auditory part of the brain in ferrets was able to learn to interpret visual inputs\cite{roe1992visual}{}. Our knowledge about necessary ingredients of such an algorithm is shaped by neuroscientific discoveries, empirical evaluation of effectiveness of algorithms, metacognition and observations. Some of the points below may be considered as very general assumptions for reverse-engineering this general-purpose learning algorithm.
\section{Ingredients}
\subsection{Unsupervised}
In real world, almost all data is unlabeled. Although, nobody kwons the precise rules used by the human brain for learning, one can assume that we learn mostly in an unsupervised way. Specifically, when a newborn learns about the world and how different objects interact, there might not even be a way to provide supervised signal to him/her, because the appropriate sensory representations (i.e. visual, auditory) need to be developed first.  Another piece of evidence against supervised learning may be obtained by simple calculation: assuming that there are approximately $10^{14}$ synapses and $10^9$ seconds of human lifetime, there is enough capacity to store all memories at the rate of $10^5$ bits/second\cite{Schmidhuber:07alt}{}. Therefore it seems reasonable that the brain learns the model of the world directly from the environment. This motivates the hypothesis of predominance of unsupervised learning, since the only way of acquiring so much information is by absorbing data from perceptual inputs\cite{hintonlecture}{}. Even when a teacher is present, most learning must be done by learning associations between events without supervision. Unsupervised learning has been researched extensively and was found to be closely connected to the process of entropy-maximization, regularization and compression\cite{DBLP:journals/corr/abs-1206-5538, MacKay:itp, hinton1999unsupervised, 888}{}. This means that through evolution, our brains have adapted to act as data compactors. In particular, the goal of unsupervised learning might be to find codes which disentangle input sources and describe the original information in a less redundant or interpretable way\cite{888}{}, by throwing out as much data as possible out without losing information. An example of this operation has been observed in the visual cortex\cite{wiesel:1959} (but might even happen as early as in the retina) which learns patterns appearing in the natural environment and assigns high probability to those patterns\cite{hyvarinen2009natural}{}. In contrast,  the cortex assigns low probability to random combinations. The real world data is said to lie near a non-linear manifold\cite{bengio2004discovering} within the higher-dimensional space, where the manifold shape is defined by the data probability distribution. Clustering is then equivalent to learning those manifolds and being able to separate them well enough for a given task.

\subsection{Compositional}
Humans learn concepts in a sequential order, first making sense of simple patterns and representing more complex ones in terms of those previously learned abstractions. The ability to read might serve as an example. First we learn to see, we recognize pen strokes, then letters, then words and then we are able to understand complex sentences. In contrast, the non-compositional approach would be to attempt to read straight from ink patterns on a piece of paper. The brain might have adapted this way to reflect the fact that the world is inherently hierarchical. And this observation also inspired the deep learning movement, which used the hierarchical approach to model real world data,  achieving unprecedented performance on many tasks. The way that deep learning algorithms automatically compose multiple layers of representations of data gives rise to models, which yield increasingly abstract associations between concepts (hence the other names used for deep learning algorithms: representation learning\cite{Barlow:89review, Bengio:2013:RLR:2498740.2498889, bengio2013deep, deng2014deep, erhan2010does} and feature learning\cite{DBLP:journals/corr/abs-1206-5538}{} among others). The main distinction between the the \emph{deep} approach and previous generation of machine learning is that the structure in the data should be discovered automatically by a general-purpose learning procedure, without the need to hand-engineer feature detectors\cite{DBLP:journals/corr/abs-1206-5538, LeCun2015}{}. This scheme agrees very well with the idea of unsupervised learning mentioned above.  In a way, abstract hierarchical representations might be a natural by-products of data compression\cite{888}{}. Given the theoretical and empirical evidence in favor of the deep representation learning, one could formulate a requirement for any type of brain-like architecture to be deep, containing many hierarchical levels.
\subsection{Sparse and Distributed}
The existence of cortical columns in the neocortex has been linked to the functional importance of such an arrangement. Each column typically responds to a sensory stimulus representing a certain body part or region of sound or vision, so that all cells belonging to that cell are excited simultaneously, therefore acting as a feature detector. At the same time, a column which is active (receiving strong input signal and spikes) will prohibit other nearby columns from becoming active. This lateral inhibition mechanism leads to sparse activity patterns. The fact that only strongly active columns will not be inhibited forces the learned patterns to be as invariant as possible, giving rise to independent \emph{feature detectors} in the cortex\cite{Bell97the`independent}{}. As one might have been expect, these sparse distributed representations in the brain (SDRs) are not coincidental, since they possess important properties from an information-theoretic perspective. The \emph{distributed} is important in order to disentangle underlying causes of variation (i.e. melody, instrument, pitch, loudness), while sparsity affects other elements of learning good features. It has been proven that given certain sparsity, a signal may be correctly reconstructed even with fewer samples than the sampling theorem requires\cite{citeulike:2688127, Donoho:2006:CS:2263438.2272089}{}.

\begin{figure}[!htbp]
\vskip 0.2in
\begin{center}
\centerline{\includegraphics[width=0.95\columnwidth]{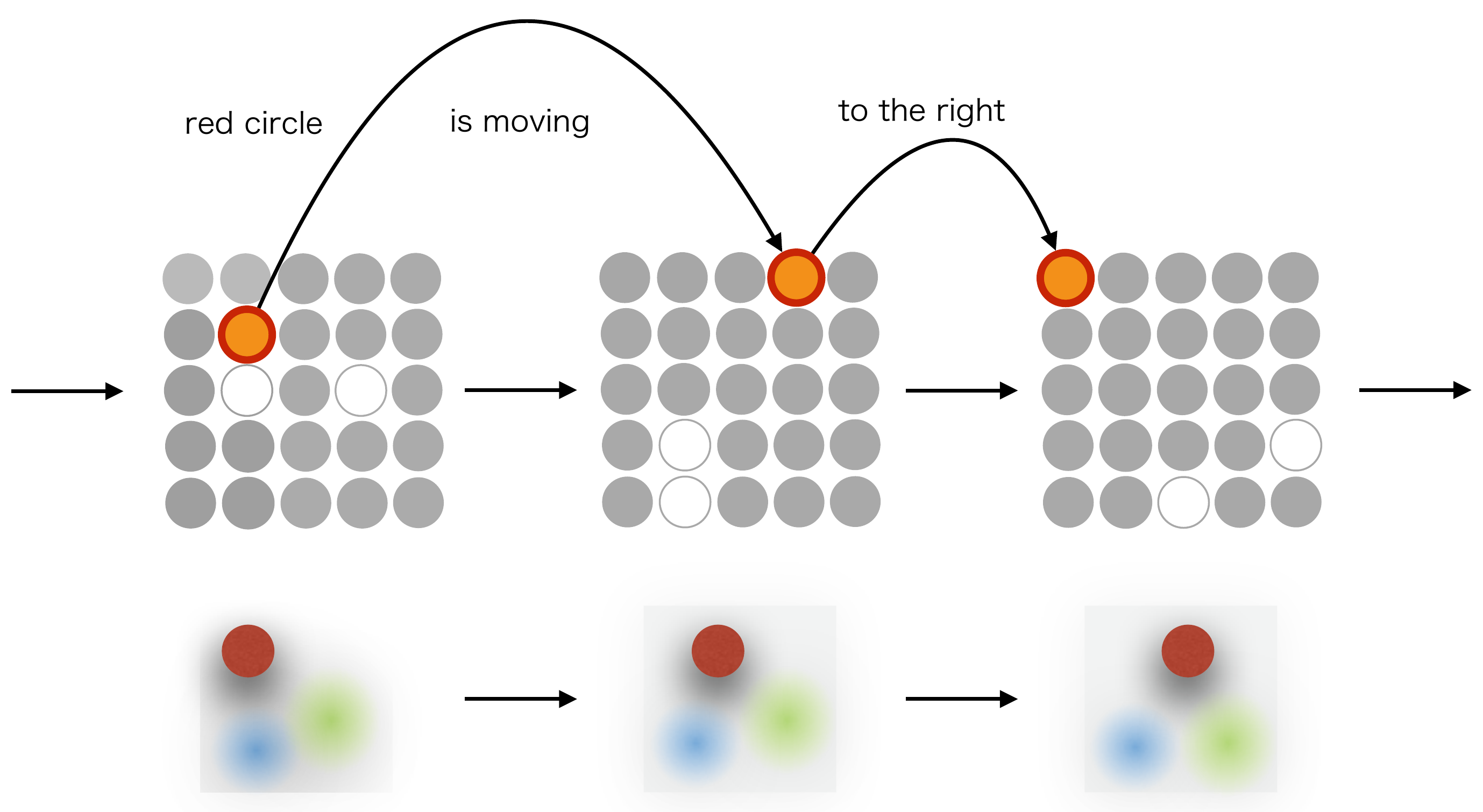}}
\caption{Efficient learning of SDRs; Sparse Distributed Representations (SDRs) simplify learning temporal dependencies; provide a mechanism for generalization and out-of-domain prediction}
\label{sdrs2}
\end{center}
\vskip -0.2in
\end{figure} 

Ever since the discovery of selective features detectors such as edge detectors and center-surround receptive fields in V1 by Hubel and Wiesel in 1959\cite{wiesel:1959}{}, learning biologically plausible sparse distributed representations of input patterns has been a hot research topic\cite{Barlow:89, Foldiak:95, Kanerva:1988:SDM:534853, Olshausen97sparsecoding}{}. It has been shown that SDRs can be significantly more noise-resistant than dense representations\cite{1503.07469}{}. Another important property of distributed representations which has been appreciated is that the number of distinguishable regions scales exponentially with the number of parameters used to describe it. This is not true for non-distributed representations. That is, sparse distributed representations are combinatorially much more expressive. Given this observation, it is simple to see that from the discriminative point of view or higher levels of abstractions, SDRs will be a preferred way of representing inputs, since the learning procedure produces a form which preserves as much information as possible while making code as short/simple as possible (also it corresponds to finding minimum-entropy codes\cite{Barlow:89, hyvarinen2001}{}). This is in-line with the Occam's Razor or Minimum Description Length (MDL) rules which postulate that simple solutions should be chosen over more complex ones\cite{Solomonoff:64, Rissanen:78}{}. This allows for manipulating sparse representations throughout the large network and simplifies learning higher level concepts (see \emph{dimensionality reduction\cite{HinSal06, saul2003}}, \emph{redundancy reduction\cite{Li:2008:IKC:1478784, doi:10.1080/net.12.3.241.253}}).

\subsection{Objectiveless}
The \emph{Chinese Room} argument\cite{searle1984minds}{} which states that learning to improve some performance measure on a given task does not necessarily lead to improving understanding of task itself. In context of supervised learning, this is not an issue, since we clearly care only about this performance measure. However, when unsupervised learning is considered, the desired outcome would be to learn transferrable concepts. It could be even hypothesized, that by following the gradient of the objective function, one may prohibit the learning procedure from discovering the unknown state-space or that progress in learning is not equivalent with being close to the objective. One hypothesis is that having an objective\cite{DBLP:books/sp/StanleyL15}{} is the problem itself. Clearly, the learning algorithm should have a goal, which might be defined very broadly such as the theory of curiosity, creativity and beauty described by J. Schmidhuber\cite{Schmidhuber:07alt}{}.

\subsection{Scalable}
In a such large network as the human brain it might be computationally efficient to separate local learning (gray matter) from adjusting higher level connections between layers/regions (white matter). This functional distinction would reflect the structural hierarchy that is so predominant in \emph{deep learning} methods described before and the real world. Biological, technological, social, transportation and other types of real-world networks are neither completely random nor definitely regular. 
\begin{figure}[!htbp]
\vskip 0.2in
\begin{center}
\centerline{\includegraphics[width=0.4\columnwidth]{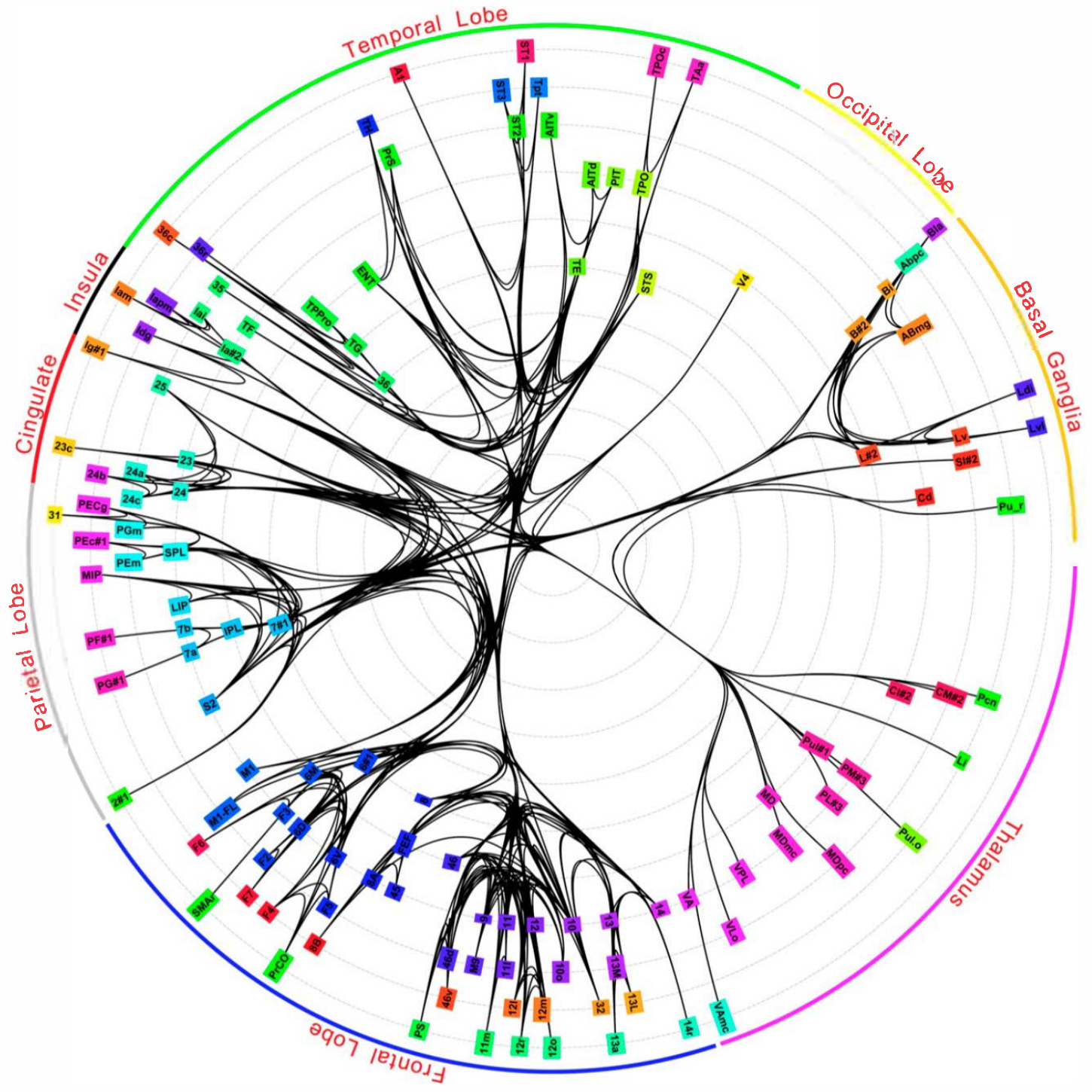}}
\caption{An example of a small world network: each edge encodes the presence of long-distance connection between corresponding regions in a macaque brain. Figure borrowed from [\citen{modha2010network}] }
\label{small}
\end{center}
\vskip -0.2in
\end{figure} 
Instead, their topology lies somewhere in between. Such so-called small world networks\cite{watts1998cds} may be nature's solution to a hierarchical structure allowing for separate parallel local and global updates of synapses, scalability and unsupervised learning at the lower levels with more goal-oriented fine-tuning in higher regions. Study of the neocortex reveals the presence of small world networks, where columnar organization reflects the local connectivity of the cerebral cortex.The brain is an inherently parallel machine, without a separate instruction-issuing and memory storage areas. Instead, all parts of the neocortex participate in both. This is a very big difference when compared to the von-Neumann architecture describing majority of computing systems are organized. The main bottleneck current systems concerns data movement, which implies additional bandwidth, power and latency requirements. CPUs are typically optimized for serial tasks, mitigating the negative effects of such an architecture by deep cache hierarchy, but losing when parallelism is involved. GPUs have more brain-like layout, with more \emph{equal} processing units, each having some private memory, so that they can actually operate in parallel without colliding. However, the problem of moving the data still exists, either between CPU and GPU or inside in the GPU. The same problem persists. In fact, it is quite easy to show, that it is virtually impossible to achieve the peak performance of those processors, because the data cannot be fed fast enough. Moreover, the data transfers are the major energy consumption factors on parallel GPU-like devices\cite{Villa:2014:SPW:2683593.2683684}{}. Therefore, a more radical approach may be needed in order to improve the performance significantly. The von-Neumann architecture needs to be changed into one where memory itself can compute. Some hardware which allows such a functionality has already appeared\cite{dlugosch2014efficient}{}. The concept of in-place processing assumes however, that a different approach is also needed when thinking about algorithms. This process of communication-aware algorithm design has already started with the advent of multi-core CPUs, GPUs and FPGAs. The next step is to design communication-less algorithms\cite{Baboulin201217}{}. This is an ongoing effort in supercomputing community, where it has been noticed, that no significant progress can be made without reducing information transfer-overhead.  

\section{Functional Ingredients}
Given some low-level properties of the learning algorithm, what should be the overall goal of learning and what should the learning path look like? What kind of behavior would be considered as a stepping stone towards machine intelligence and if so, is there a way to describe it in a precise way? Even the very basic question of what it means for a machine or an algorithm be intelligent needs clarification. According to some, goal-directed behavior is considered the essence of intelligence\cite{Russell:2003:AIM:773294}{}. However, this implies that the necessary and sufficient condition of intelligent behavior is rationality and this paper questions this statement. Humans are often very far from being rational. Creativity does not fall into this definition and risk-taking might not be rational, yet both are essential for innovation. Therefore, far more appealing theories of universal intelligence are those with broader priors, such as the theory of curiosity, creativity and beauty described by J. Schmidhuber\cite{Schmidhuber:07alt}{}. The previous section introduces problems which may arise from objective based learning, that is the \emph{Chinese Room} argument, when all the algorithm attempts to map inputs to outputs without any motivation to learn anything beyond the task given. An intelligent algorithm (strong AI\cite{searle1984minds}{}, among other names) should be able to reveal hidden knowledge which might not even be discoverable to humans. This section describes functional ingredients of any learning procedure which would not violate the generality assumption.

\subsection{Compression}
Learning may be likened to a formal information-theory based concept of information compression. Assuming that the goal is to build more compact and more useful representations of the environment(such as finding minimum entropy codes\cite{journals/neco/BarlowKM89}{}), this interpretation relates to representation learning and analogy building compression scheme\cite{DBLP:journals/corr/abs-1108-1169} of the neocortex. One way of looking at this task is considering a general artificial intelligence as a general purpose compressor, one which is able to discover the probability distribution of any source\cite{MacKay:itp}{}. However, the \emph{No Free Lunch Theorem}\cite{Wolpert:1997:NFL:2221336.2221408} states that no completely general-purpose learning algorithm can exist. In other words, for any given compressor, there exists a data distribution on which it will perform poorly. This implies that there must exist some restrictions on the class of problems such a learning system can address as well. The previous section already mentioned a few of them, which are fortunately very general and plausible such as the \emph{smoothness prior} or \emph{depth prior}{} (also see [\citen{Bengio+chapter2007}] for a more complete list of sensible assumptions). 

\subsection{Prediction}
Whereas the smoothness prior may be considered as a type of spatial coherence, the assumption that the world is mostly predictable corresponds to temporal or more generally spatiotemporal coherence. This is probably the most important ingredient of a general-purpose learning procedure. Such an assumption states that things which close in time are close in space and vice versa. A purely spatial analogy is huge image space yet only a tiny fraction of possible real images\cite{hyvarinen2009natural}{}. The same is true for spatiotemporal patterns. The assumption that a  sequence of spatial patterns is coherent restricts the spectrum of future spatial states which are likely.
Occam's Razor rule or the MDL principle\cite{Solomonoff:64, Rissanen:78}{} state that simple solutions should be favored over more complex ones. Therefore, learning better representations should be a goal itself, even without any other objective. If it is assumed that no task is given a priori, the best we can do is just to observe and learn to predict. One of the first working examples (and a proof of concept) is the principle of history compression employed in the recurrent architecture proposed by J. Schidmuber\cite{schmidhuber1992}{}.

\subsection{Understanding}
The ability to predict is equivalent to understanding, since at any given moment, a cause and prediction could be inferred from given state context. Therefore, learning to predict may be a more general requirement of an intelligent behavior. In fact, it has been postulated\cite{Hawkins:2004:INT:993636}{} that all the brain does is constantly predict future states, compare those predictions with sensory inputs, and readjust accordingly. While this might seem to be equivalent to backpropagating the error through the entire network, however from the biological perspective, the prediction/expectation readjustment of neurons is most likely operating locally.

\subsection{Sensorimotor}
\begin{figure}[!htbp]
\vskip 0.2in
\begin{center}
\centerline{\includegraphics[width=0.95\columnwidth]{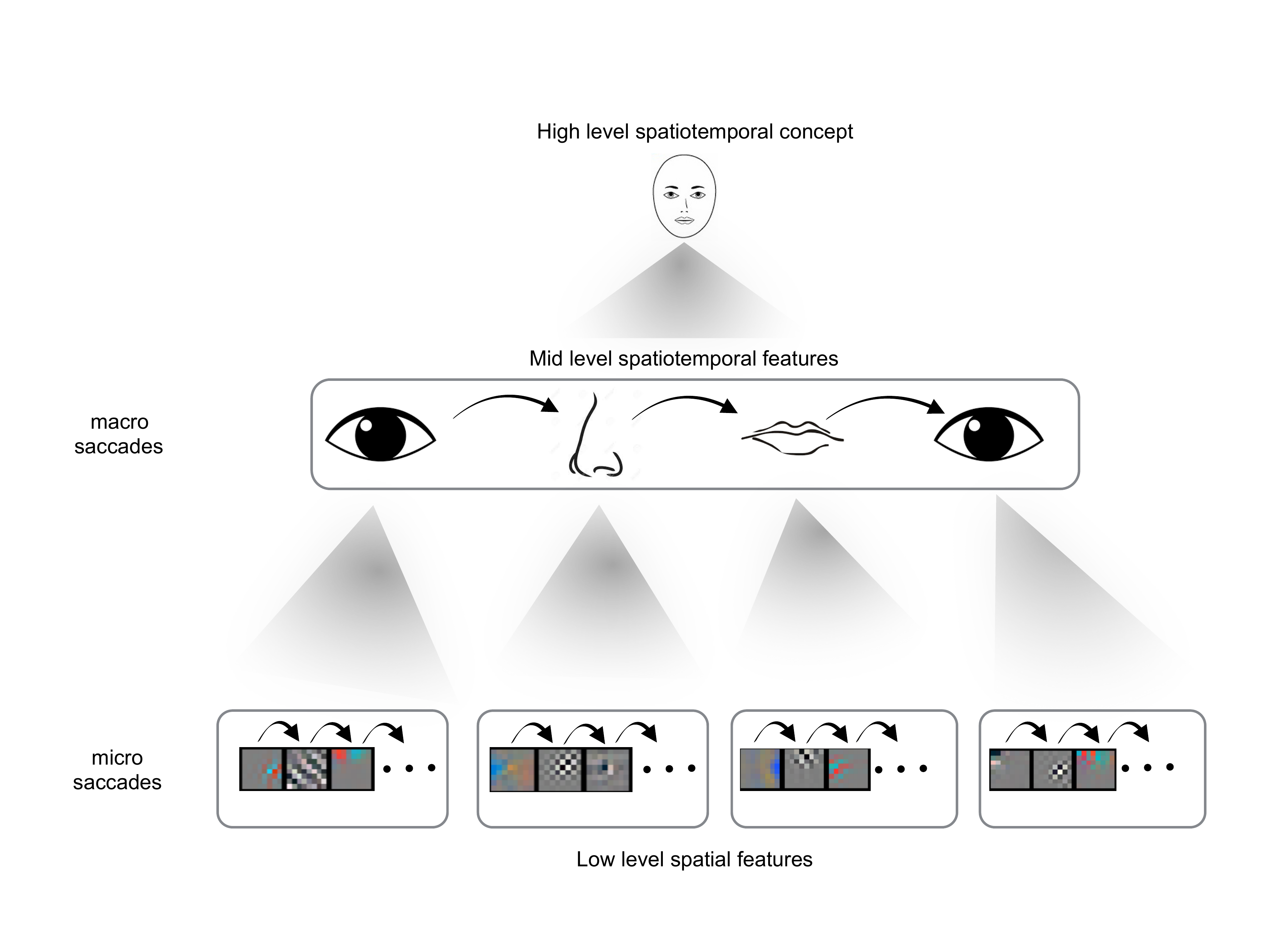}}
\caption{Face as an example of a spatiotemporal concept, micro-saccades are sequences of low-level spatial patterns in the fovea, they can be pooled temporally into a mid-level concept of an eye, or nose; macro-saccades are more task-oriented movement - moving between nose, eyes, mouth}
\label{face}
\end{center}
\vskip -0.2in
\end{figure}

Scientists have demonstrated that the brain predicts consequences of our eye movements based on what we see next. The findings have implications for understanding human attention and applications to robotics. Despite the fact that, in practice, no experienced can be perceived twice, human brains are able to form a stable representation of abstract concepts and make accurate predictions despite changes in context. Such mental representations help explain the rapid eye movements known as saccades. Our eyes move rapidly approximately three times a second in order to capture new visual information. With each jump a new image falls onto the retina. However, we do not experience this quickly-changing sequence of images, instead, we see a stable image (Fig. \ref{face}). The brain uses such a mechanism in order to redirect attention, since only approximately 1$^{\circ}$ of the retina provides sharp image (fovea). This operation has been extensively researched from the neuroscientific perspective as it provides one of few visible brain activities\cite{Rolfs2011, Kowler20111457}. Sensorimotor connections are needed in order to know which changes in the image do not result from internal eye movement and which do not. One hypothesis is that the basic repeating functional unit of the neocortex is a sensorimotor model\cite{hawkins2015neurons}{}, that is every part of the brain performs both sensory and motor processing to some extent. Complex cells in V2 visual cortex which are invariant to small changes of inputs patterns\cite{lee2007sparse}{} might be mapped purely spatially or may represent a spatiotemporal patterns (i.e. invariant representation given an action). Other experiments support the claim, showing a similar mechanism operating on different type of sensory inputs\cite{Krieger:2015:SIW:2838985, Diamond2008}.

\subsection{Spatiotemporal Invariance}
\begin{figure}[!htbp]
\vskip 0.2in
\begin{center}
\centerline{\includegraphics[width=0.7\columnwidth]{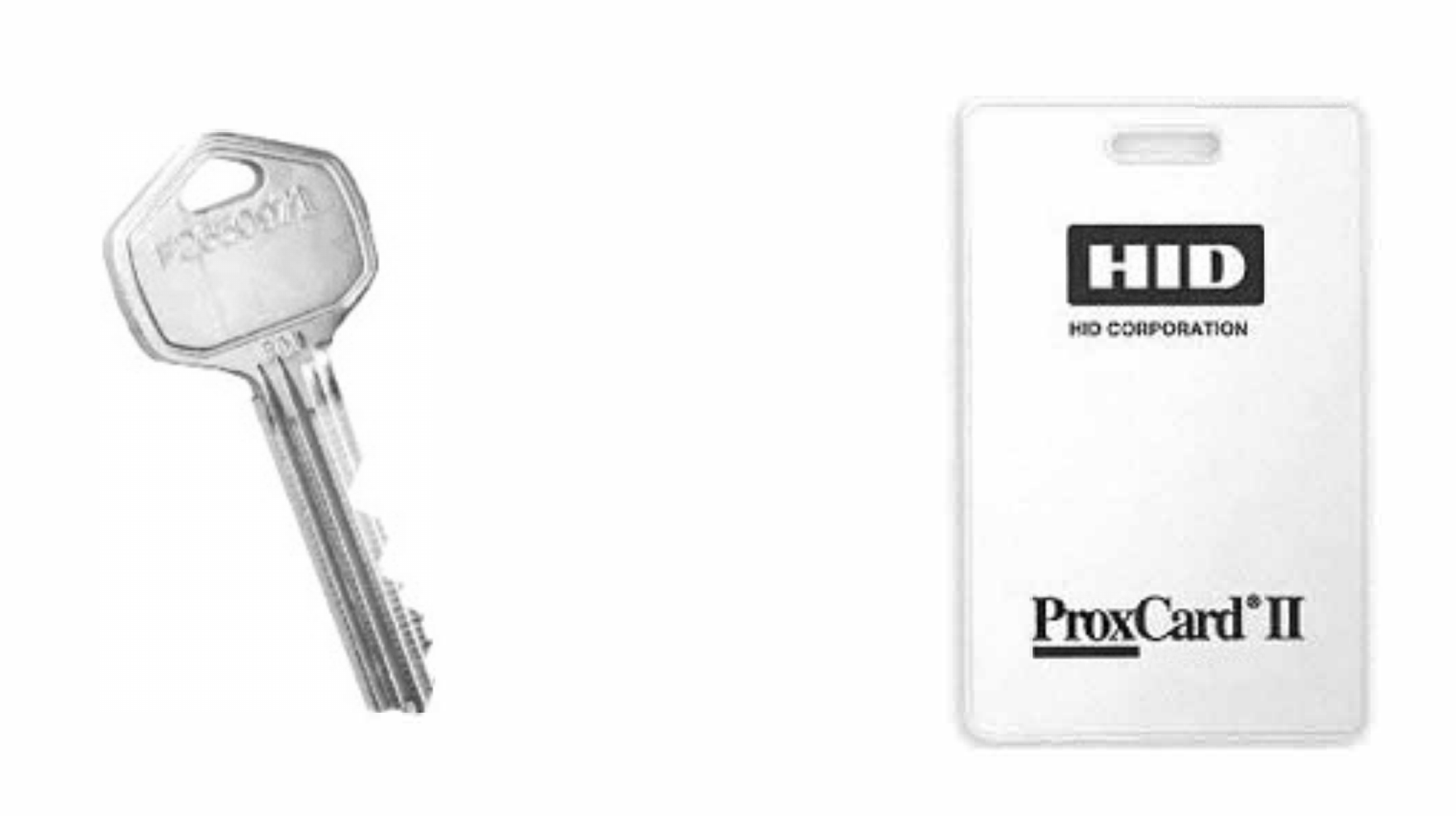}}
\caption{An example of a spatiotemporal concept}
\label{keys}
\end{center}
\vskip -0.2in
\end{figure}

Thinking about motor command in a more abstract way, it is possible to show that in order to disambiguate multiple predictions. one needs to \emph{inject additional context}. This paper assumes that predictions are associated with some uncertainty\cite{ref1, 10.1371/journal.pcbi.1004305} as in the bayesian approach and that instead of assuming a single point prediction, the distribution is highly multimodal. Additional context is equivalent to integrating evidence which makes predictions more specific. The need for abstract spatiotemporal concepts can be illustrated with a simple example. Given two images as in Fig. \ref{keys}, it is obvious that classification based on purely spatial aspect of a pattern can be inadequate. A much more natural way of grouping these two objects is by their function, which requires an ability to \emph{imagine} whether a particular object can be used in a certain way (in this case, to open a door). The same considerations apply to other objects, such as chairs. It is much more natural to learn these concepts as spatiotemporal ideas rather than predominantly depend on spatial appearance. When considering the ability to \emph{imagine/dream/hallucinate}, then widespread implementation of sensorimotor functionality in the brain is not very surprising. The concept of manipulating a compact spatiotemporal thought might be necessary from the reasoning perspective\cite{bottou-mlj-2013}{} or transfer learning, as majority of the analogies we make are temporal in nature. The importance of learning transformations in the real-world has been recognized in the research community\cite{memisevic2010, Boulanger-et-al-ICML2012, Sutskever, sutskever2008, WisSej2002, Elman90findingstructure, 888}{}, but still needs more attention.

\subsection{Context update/pattern completion}
\begin{figure}[!htbp]
\vskip 0.2in
\begin{center}
\centerline{\includegraphics[width=0.8\columnwidth]{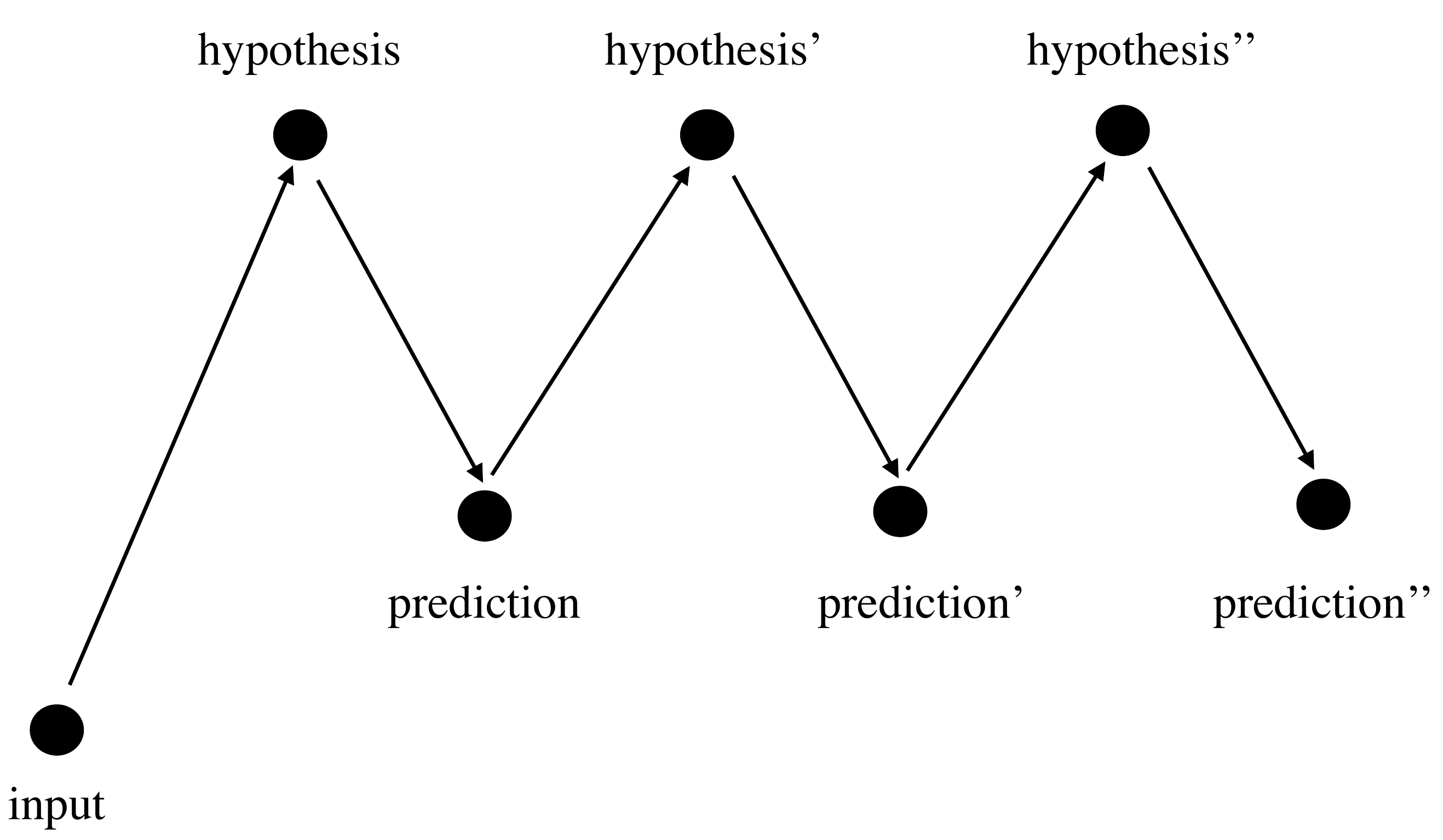}}
\caption{Illustration of iterative context update, every prediction changes the context slightly and vice-versa}
\label{f:rerun}
\end{center}
\vskip -0.2in
\end{figure}

The last functional component postulated by this paper is a continuous (in theory) loop between bottom-up predictions and top-down context. The hypothesis is that such interconnectedness enables perceptual completion, where higher layers make hypotheses about the inferences coming from the lower layers and then predictions are iteratively refined based on those hypotheses. This may be likened to working memory theory, where non-episodic memories are being held (not involving hippocampus). An analogy of this is Expectation Maximization or the learning procedure commonly used in Boltzmann Machines, where a samples are obtained iteratively by alternating between unit activations on two connected layers\cite{series/lncs/Hinton12, resnik2010gibbs}{} (see Fig. \ref{f:rerun}). A real-world analogy of this process is solving a crossword or a sudoku puzzle or filling in missing words in a sentence. Such problems may require iterative solution refinement procedure.

\section*{Acknowledgments} 
Partial support for this work was provided by the Defense Advanced Research Projects Agency (DARPA). I would like to thank members of machine intelligence group at IBM Research and Numenta for their suggestions and many interesting discussions.

\bibliographystyle{splncs03}
\bibliography{typeinst}

\begin{thebibliography}{10}
\providecommand{\url}[1]{\texttt{#1}}
\providecommand{\urlprefix}{URL }

\bibitem{1503.07469}
Ahmad, S., Hawkins, J.: Properties of sparse distributed representations and
  their application to hierarchical temporal memory (2015)

\bibitem{Baboulin201217}
Baboulin, M., Donfack, S., Dongarra, J., Grigori, L., Rémy, A., Tomov, S.: A
  class of communication-avoiding algorithms for solving general dense linear
  systems on cpu/gpu parallel machines. Procedia Computer Science  9,  17 -- 26
  (2012),
  \url{http://www.sciencedirect.com/science/article/pii/S187705091200124X},
  proceedings of the International Conference on Computational Science,
  \{ICCS\} 2012

\bibitem{doi:10.1080/net.12.3.241.253}
Barlow, H.: Redundancy reduction revisited. Network: Computation in Neural
  Systems  12(3),  241--253 (2001),
  \url{http://dx.doi.org/10.1080/net.12.3.241.253}, pMID: 11563528

\bibitem{Barlow:89review}
Barlow, H.B.: Unsupervised learning. Neural Computation  1(3),  295--311 (1989)

\bibitem{Barlow:89}
Barlow, H.B., Kaushal, T.P., Mitchison, G.J.: Finding minimum entropy codes.
  Neural Computation  1(3),  412--423 (1989)

\bibitem{journals/neco/BarlowKM89}
Barlow, H.B., Kaushal, T.P., Mitchison, G.J.: Finding minimum entropy codes.
  Neural Computation  1(3),  412--423 (1989),
  \url{http://dblp.uni-trier.de/db/journals/neco/neco1.html#BarlowKM89}

\bibitem{Bell97the`independent}
Bell, A.J., Sejnowski, T.J.: The `independent components' of natural scenes are
  edge filters. VISION RESEARCH  37,  3327--3338 (1997)

\bibitem{Bengio-2009}
Bengio, Y.: Learning deep architectures for {AI}. Foundations and Trends in
  Machine Learning  2(1),  1--127 (2009), also published as a book. Now
  Publishers, 2009.

\bibitem{bengio2013deep}
Bengio, Y.: Deep learning of representations: Looking forward. In: Statistical
  Language and Speech Processing, pp. 1--37. Springer (2013)

\bibitem{Bengio:2013:RLR:2498740.2498889}
Bengio, Y., Courville, A., Vincent, P.: Representation learning: A review and
  new perspectives. IEEE Trans. Pattern Anal. Mach. Intell.  35(8),  1798--1828
  (Aug 2013), \url{http://dx.doi.org/10.1109/TPAMI.2013.50}

\bibitem{DBLP:journals/corr/abs-1206-5538}
Bengio, Y., Courville, A.C., Vincent, P.: Unsupervised feature learning and
  deep learning: {A} review and new perspectives. CoRR  abs/1206.5538 (2012),
  \url{http://arxiv.org/abs/1206.5538}

\bibitem{Bengio+chapter2007}
Bengio, Y., {LeCun}, Y.: Scaling learning algorithms towards {AI}. In: Bottou,
  L., Chapelle, O., DeCoste, D., Weston, J. (eds.) Large Scale Kernel Machines.
  MIT Press (2007),
  \url{http://www.iro.umontreal.ca/~lisa/pointeurs/bengio+lecun_chapter2007.pdf}

\bibitem{bengio2004discovering}
Bengio, Y., Monperrus, M.: Discovering shared structure in manifold learning
  (2004)

\bibitem{bottou-mlj-2013}
Bottou, L.: From machine learning to machine reasoning: an essay. Machine
  Learning  94,  133--149 (January 2014),
  \url{http://leon.bottou.org/papers/bottou-mlj-2013}

\bibitem{Boulanger-et-al-ICML2012}
Boulanger-Lewandowski, N., Bengio, Y., Vincent, P.: Modeling temporal
  dependencies in high-dimensional sequences: Application to polyphonic music
  generation and transcription. In: Proceedings of the Twenty-nine
  International Conference on Machine Learning (ICML'12). ACM (2012),
  \url{http://icml.cc/discuss/2012/590.html}

\bibitem{citeulike:2688127}
Cand\`{e}s, E.J., Romberg, J.K., Tao, T.: {Stable signal recovery from
  incomplete and inaccurate measurements}. Comm. Pure Appl. Math.  59(8),
  1207--1223 (Aug 2006), \url{http://dx.doi.org/10.1002/cpa.20124}

\bibitem{deng2014deep}
Deng, L., Yu, D.: Deep learning: methods and applications. Foundations and
  Trends in Signal Processing  7(3--4),  197--387 (2014)

\bibitem{Diamond2008}
Diamond, M.E., von Heimendahl, M., Knutsen, P.M., Kleinfeld, D., Ahissar, E.:
  'where' and 'what' in the whisker sensorimotor system. Nat Rev Neurosci
  9(8),  601--612 (Aug 2008), \url{http://dx.doi.org/10.1038/nrn2411}

\bibitem{dlugosch2014efficient}
Dlugosch, P., Brown, D., Glendenning, P., Leventhal, M., Noyes, H.: An
  efficient and scalable semiconductor architecture for parallel automata
  processing. Parallel and Distributed Systems, IEEE Transactions on  25(12),
  3088--3098 (2014)

\bibitem{domingos2015master}
Domingos, P.: The Master Algorithm: How the Quest for the Ultimate Learning
  Machine Will Remake Our World. Penguin Books Limited (2015),
  \url{https://books.google.com/books?id=pjRkCQAAQBAJ}

\bibitem{Donoho:2006:CS:2263438.2272089}
Donoho, D.L.: Compressed sensing. IEEE Trans. Inf. Theor.  52(4),  1289--1306
  (Apr 2006), \url{http://dx.doi.org/10.1109/TIT.2006.871582}

\bibitem{Elman90findingstructure}
Elman, J.L.: Finding structure in time. COGNITIVE SCIENCE  14(2),  179--211
  (1990)

\bibitem{erhan2010does}
Erhan, D., Bengio, Y., Courville, A., Manzagol, P.A., Vincent, P., Bengio, S.:
  Why does unsupervised pre-training help deep learning? The Journal of Machine
  Learning Research  11,  625--660 (2010)

\bibitem{Foldiak:95}
F\"{o}ldi\'{a}k, P., Young, M.P.: Sparse coding in the primate cortex. In:
  Arbib, M.A. (ed.) The Handbook of Brain Theory and Neural Networks. pp.
  895--898. The MIT Press (1995)

\bibitem{Goodfellow-et-al-2015-Book}
Goodfellow, I., Courville, A., Bengio, Y.: Deep learning (2015),
  \url{http://goodfeli.github.io/dlbook/}, book in preparation for MIT Press

\bibitem{Graves:09tpami}
Graves, A., Liwicki, M., Fernandez, S., Bertolami, R., Bunke, H., Schmidhuber,
  J.: A novel connectionist system for improved unconstrained handwriting
  recognition. IEEE Transactions on Pattern Analysis and Machine Intelligence
  31(5) (2009)

\bibitem{graves2014}
Graves, A., Jaitly, N.: Towards end-to-end speech recognition with recurrent
  neural networks. In: Proc. 31st International Conference on Machine Learning
  (ICML). pp. 1764--1772 (2014)

\bibitem{DBLP:journals/corr/abs-1108-1169}
Gregor, K., LeCun, Y.: Learning representations by maximizing compression. CoRR
   abs/1108.1169 (2011), \url{http://arxiv.org/abs/1108.1169}

\bibitem{hawkins2015neurons}
Hawkins, J., Ahmad, S.: Why neurons have thousands of synapses, a theory of
  sequence memory in neocortex. arXiv preprint arXiv:1511.00083  (2015)

\bibitem{Hawkins:2004:INT:993636}
Hawkins, J., Blakeslee, S.: On Intelligence. Times Books (2004)

\bibitem{HintonSejnowski:86}
Hinton, G.E., Sejnowski, T.E.: Learning and relearning in {Boltzmann} machines.
  In: Parallel Distributed Processing, vol.~1, pp. 282--317. MIT Press (1986)

\bibitem{hinton1999unsupervised}
Hinton, G., Sejnowski, T.: Unsupervised Learning: Foundations of Neural
  Computation. A Bradford Book, MCGRAW HILL BOOK Company (1999),
  \url{https://books.google.com/books?id=yj04Y0lje4cC}

\bibitem{HinSal06}
Hinton, G., Salakhutdinov, R.: Reducing the dimensionality of data with neural
  networks. Science  313(5786),  504--507 (2006)

\bibitem{hintonlecture}
Hinton, G.E.: {Learning Representations by Unlearning Beliefs}.
  \url{http://www.ircs.upenn.edu/pinkel/lectures/hinton/Hinton_PinkelTranscription_2003.pdf}
  (2003), [Online; accessed 23-November-2015]

\bibitem{series/lncs/Hinton12}
Hinton, G.E.: A practical guide to training restricted boltzmann machines. In:
  Montavon, G., Orr, G.B., Müller, K.R. (eds.) Neural Networks: Tricks of the
  Trade (2nd ed.), Lecture Notes in Computer Science, vol. 7700, pp. 599--619.
  Springer (2012),
  \url{http://dblp.uni-trier.de/db/series/lncs/lncs7700.html#Hinton12}

\bibitem{hyvarinen2009natural}
Hyv{\"a}rinen, A., Hurri, J., Hoyer, P.O.: Natural Image Statistics: A
  Probabilistic Approach to Early Computational Vision., vol.~39. Springer
  Science \& Business Media (2009)

\bibitem{hyvarinen2001}
Hyv{\"a}rinen, A., Karhunen, J., Oja, E.: Independent component analysis. John
  Wiley \& Sons (2001)

\bibitem{Kanerva:1988:SDM:534853}
Kanerva, P.: Sparse Distributed Memory. MIT Press, Cambridge, MA, USA (1988)

\bibitem{Kowler20111457}
Kowler, E.: Eye movements: The past 25 years. Vision Research  51(13),  1457 --
  1483 (2011),
  \url{http://www.sciencedirect.com/science/article/pii/S0042698910005924},
  vision Research 50th Anniversary Issue: Part 2

\bibitem{Krieger:2015:SIW:2838985}
Krieger, P., Groh, A.: Sensorimotor Integration in the Whisker System. Springer
  Publishing Company, Incorporated, 1st edn. (2015)

\bibitem{NIPS2012_4824}
Krizhevsky, A., Sutskever, I., Hinton, G.E.: Imagenet classification with deep
  convolutional neural networks. In: Pereira, F., Burges, C., Bottou, L.,
  Weinberger, K. (eds.) Advances in Neural Information Processing Systems 25,
  pp. 1097--1105. Curran Associates, Inc. (2012),
  \url{http://papers.nips.cc/paper/4824-imagenet-classification-with-deep-convolutional-neural-networks.pdf}

\bibitem{kurzweil2012create}
Kurzweil, R.: How to Create a Mind: The Secret of Human Thought Revealed.
  Penguin Publishing Group (2012),
  \url{https://books.google.com/books?id=FCcXiBPurdEC}

\bibitem{lecuncvpr}
LeCun, Y.: {What's Wrong With Deep Learning?}
  \url{http://www.pamitc.org/cvpr15/files/lecun-20150610-cvpr-keynote.pdf}
  (2015), [Online; accessed 20-November-2015]

\bibitem{LeCun2015}
LeCun, Y., Bengio, Y., Hinton, G.: Deep learning. Nature  521(7553),  436--444
  (May 2015), \url{http://dx.doi.org/10.1038/nature14539}, insight

\bibitem{lee2007sparse}
Lee, H., Ekanadham, C., Ng, A.Y.: Sparse deep belief net model for visual area
  {V2}. In: Advances in Neural Information Processing Systems (NIPS). vol.~7,
  pp. 873--880 (2007)

\bibitem{Li:2008:IKC:1478784}
Li, M., Vitnyi, P.M.: An Introduction to Kolmogorov Complexity and Its
  Applications. Springer Publishing Company, Incorporated, 3 edn. (2008)

\bibitem{MacKay:itp}
MacKay, D.J.C.: Information Theory, Inference, and Learning Algorithms.
  Cambridge University Press (2003), \url{http://www.cambridge.org/0521642981},
  {}

\bibitem{memisevic2010}
Memisevic, R., Hinton, G.E.: Learning to represent spatial transformations with
  factored higher-order {Boltzmann} machines. Neural Computation  22(6),
  1473--1492 (2010)

\bibitem{10.1371/journal.pcbi.1004305}
Meyniel, F., Schlunegger, D., Dehaene, S.: The sense of confidence during
  probabilistic learning: A normative account. PLoS Comput Biol  11(6),
  e1004305 (06 2015), \url{http://dx.doi.org/10.1371%2Fjournal.pcbi.1004305}

\bibitem{ref1}
Meyniel, F., Sigman, M., Mainen, Z.: Confidence as bayesian probability: From
  neural origins to behavior. Neuron  88(1),  78--92 (2015/11/25 XXXX),
  \url{http://dx.doi.org/10.1016/j.neuron.2015.09.039}

\bibitem{Mnih2015}
Mnih, V., Kavukcuoglu, K., Silver, D., Rusu, A.A., Veness, J., Bellemare, M.G.,
  Graves, A., Riedmiller, M., Fidjeland, A.K., Ostrovski, G., Petersen, S.,
  Beattie, C., Sadik, A., Antonoglou, I., King, H., Kumaran, D., Wierstra, D.,
  Legg, S., Hassabis, D.: Human-level control through deep reinforcement
  learning. Nature  518(7540),  529--533 (Feb 2015),
  \url{http://dx.doi.org/10.1038/nature14236}, letter

\bibitem{modha2010network}
Modha, D.S., Singh, R.: Network architecture of the long-distance pathways in
  the macaque brain. Proceedings of the National Academy of Sciences  107(30),
  13485--13490 (2010)

\bibitem{mohamed2009}
Mohamed, A., Dahl, G.E., Hinton, G.E.: Deep belief networks for phone
  recognition. In: NIPS'22 workshop on deep learning for speech recognition
  (2009)

\bibitem{Mountcastle:1978}
Mountcastle, V.B.: An organizing principle for cerebral function: The unit
  model and the distributed system. In: Edelman, G.M., Mountcastle, V.V. (eds.)
  The Mindful Brain, pp. 7--50. MIT Press, Cambridge, MA (1978)

\bibitem{citeulike:13329708}
Ng, A.: {The Man Behind the Google Brain: Andrew Ng and the Quest for the New
  AI} (Jul 2013),
  \url{http://www.wired.com/2013/05/neuro-artificial-intelligence/},
  http://www.wired.com/2013/05/neuro-artificial-intelligence/

\bibitem{Olshausen97sparsecoding}
Olshausen, B.A., Field, D.J.: Sparse coding with an overcomplete basis set: a
  strategy employed by v1. Vision Research  37,  3311--3325 (1997)

\bibitem{resnik2010gibbs}
Resnik, P., Hardisty, E.: Gibbs sampling for the uninitiated. Tech. rep., DTIC
  Document (2010)

\bibitem{Rissanen:78}
Rissanen, J.: Modeling by shortest data description. Automatica  14,  465--471
  (1978)

\bibitem{roe1992visual}
Roe, A.W., Pallas, S.L., Kwon, Y.H., Sur, M.: Visual projections routed to the
  auditory pathway in ferrets: receptive fields of visual neurons in primary
  auditory cortex. The Journal of neuroscience  12(9),  3651--3664 (1992)

\bibitem{Rolfs2011}
Rolfs, M., Jonikaitis, D., Deubel, H., Cavanagh, P.: Predictive remapping of
  attention across eye movements. Nat Neurosci  14(2),  252--256 (Feb 2011),
  \url{http://dx.doi.org/10.1038/nn.2711}

\bibitem{Russell:2003:AIM:773294}
Russell, S.J., Norvig, P.: Artificial Intelligence: A Modern Approach. Pearson
  Education, 2 edn. (2003)

\bibitem{saul2003}
Saul, L.K., Roweis, S.T.: Think globally, fit locally: unsupervised learning of
  low dimensional manifolds. The Journal of Machine Learning Research  4,
  119--155 (2003)

\bibitem{schmidhuber1992}
Schmidhuber, J.: Learning complex, extended sequences using the principle of
  history compression. Neural Computation  4(2),  234--242 (1992)

\bibitem{Schmidhuber:07alt}
Schmidhuber, J.: Simple algorithmic principles of discovery, subjective beauty,
  selective attention, curiosity \& creativity. In: Proc. 18th Intl. Conf. on
  Algorithmic Learning Theory (ALT 2007), LNAI 4754. pp. 32--33. Springer
  (2007), joint invited lecture for {\em ALT 2007 and DS 2007}, Sendai, Japan,
  2007

\bibitem{888}
Schmidhuber, J.: Deep learning in neural networks: An overview. Neural Networks
   61,  85--117 (2015), published online 2014; based on TR arXiv:1404.7828
  [cs.NE]

\bibitem{searle1984minds}
Searle, J.: Minds, Brains, and Science. Reith lectures, Harvard University
  Press (1984), \url{https://books.google.com/books?id=yNJN-\_jznw4C}

\bibitem{Solomonoff:64}
Solomonoff, R.J.: A formal theory of inductive inference. {Part I}. Information
  and Control  7,  1--22 (1964)

\bibitem{DBLP:books/sp/StanleyL15}
Stanley, K.O., Lehman, J.: Why Greatness Cannot Be Planned - The Myth of the
  Objective. Springer (2015), \url{http://dx.doi.org/10.1007/978-3-319-15524-1}

\bibitem{Sutskever}
Sutskever, I., Hinton, G.: {Learning multilevel distributed representations for
  high-dimensional sequences}. AISTATS  (2007),
  \url{http://machinelearning.wustl.edu/mlpapers/paper\_files/AISTATS07\_SutskeverH.pdf}

\bibitem{sutskever2008}
Sutskever, I., Hinton, G.E., Taylor, G.W.: The recurrent temporal restricted
  {Boltzmann} machine. In: NIPS. vol.~21, p. 2008 (2008)

\bibitem{1409.4842}
Szegedy, C., Liu, W., Jia, Y., Sermanet, P., Reed, S., Anguelov, D., Erhan, D.,
  Vanhoucke, V., Rabinovich, A.: Going deeper with convolutions (2014)

\bibitem{Villa:2014:SPW:2683593.2683684}
Villa, O., Johnson, D.R., O'Connor, M., Bolotin, E., Nellans, D., Luitjens, J.,
  Sakharnykh, N., Wang, P., Micikevicius, P., Scudiero, A., Keckler, S.W.,
  Dally, W.J.: Scaling the power wall: A path to exascale. In: Proceedings of
  the International Conference for High Performance Computing, Networking,
  Storage and Analysis. pp. 830--841. SC '14, IEEE Press, Piscataway, NJ, USA
  (2014), \url{http://dx.doi.org/10.1109/SC.2014.73}

\bibitem{watts1998cds}
Watts, D.J., Strogatz, S.H.: {Collective dynamics of'small-world'networks.}
  Nature  393(6684),  409--10 (1998)

\bibitem{wiesel:1959}
Wiesel, D.H., Hubel, T.N.: Receptive fields of single neurones in the cat's
  striate cortex. J. Physiol.  148,  574--591 (1959)

\bibitem{WisSej2002}
Wiskott, L., Sejnowski, T.: Slow feature analysis: Unsupervised learning of
  invariances. Neural Computation  14(4),  715--770 (2002)

\bibitem{Wolpert:1997:NFL:2221336.2221408}
Wolpert, D.H., Macready, W.G.: No free lunch theorems for optimization. Trans.
  Evol. Comp  1(1),  67--82 (Apr 1997),
  \url{http://dx.doi.org/10.1109/4235.585893}

\end{thebibliography}

\end{document}